\pdfoutput=1

\documentclass{svproc}
\usepackage{url}

\usepackage{graphicx}

\usepackage{geometry}
\begin{document}
\mainmatter              
\title{EDEN: Deep Feature Distribution Pooling for Saimaa Ringed Seals Pattern Matching
}
\author{Ilia Chelak\inst{1} \inst{2}\and
Ekaterina Nepovinnykh\inst{2}
\and
Tuomas Eerola\inst{2} 
\and 
Heikki K\"alvi\"ainen\inst{2}
\and
Igor Belykh\inst{1} }

\authorrunning{I. Chelak et al.}
\tocauthor{E. Nepovinnykh, T. Eerola, and H. K\"alvi\"ainen}
\institute{
Peter the Great St. Petersburg Polytechnic University, 
Saint Petersburg, Russian Federation \\
\email{chelak.ir@edu.spbstu.ru, igor.belyh@cit.icc.spbstu.ru},
\and
Lappeenranta-Lahti University of Technology LUT, 
School of Engineering Science, 
Department of Computational Engineering, 
Computer Vision and Pattern Recognition Laboratory, 
P.O.Box 20, 53850 Lappeenranta, Finland\\
\email{firstname.lastname@lut.fi}
}

\maketitle              

\begin{abstract}
In this paper, pelage pattern matching is considered to solve the individual re-identification of the Saimaa ringed seals. Animal re-identification together with the access to large amount of image material through camera traps and crowd-sourcing provide novel possibilities for animal monitoring and conservation. We propose a novel feature pooling approach that allow aggregating the local pattern features to get a fixed size embedding vector that incorporate global features by taking into account the spatial distribution of features. This is obtained by eigen decomposition of covariances computed for probability mass functions representing feature maps. Embedding vectors can then be used to find the best match in the database of known individuals allowing animal re-identification. The results show that the proposed pooling method outperforms the existing methods on the challenging Saimaa ringed seal image data.
%
\end{abstract}
%
%
\section{Introduction}
%

Automatic camera traps (game cameras) and crowd-sourcing provide tools for collecting large volumes of image material to monitor and to study wildlife animals. This has made it possible for researchers to obtain versatile and novel knowledge about animal populations such as population size, moving patterns, and social behaviour. The analysis of massive image volumes calls for automatic solutions motivating the use of computer vision techniques. The main computer vision problem to be solved is the photo-identification, i.e., the re-identification of the individual animals based on certain visual characteristics such as fur pattern.       

\begin{figure}[ht]
		\centering
		\minipage{0.30\linewidth}
		\includegraphics[width=\linewidth]{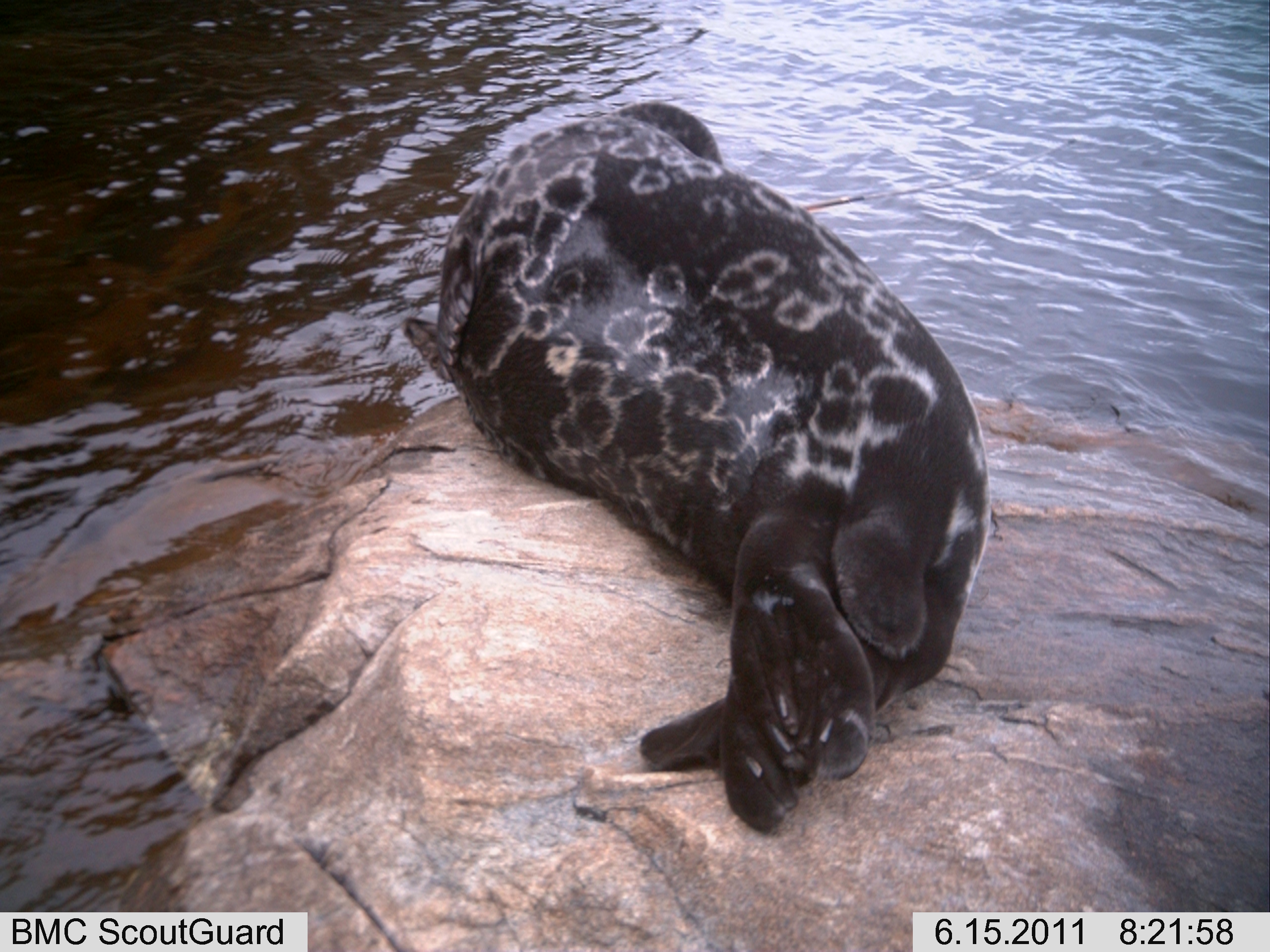}
		\endminipage\hfill
		\minipage{0.30\linewidth}
		\includegraphics[width=\linewidth]{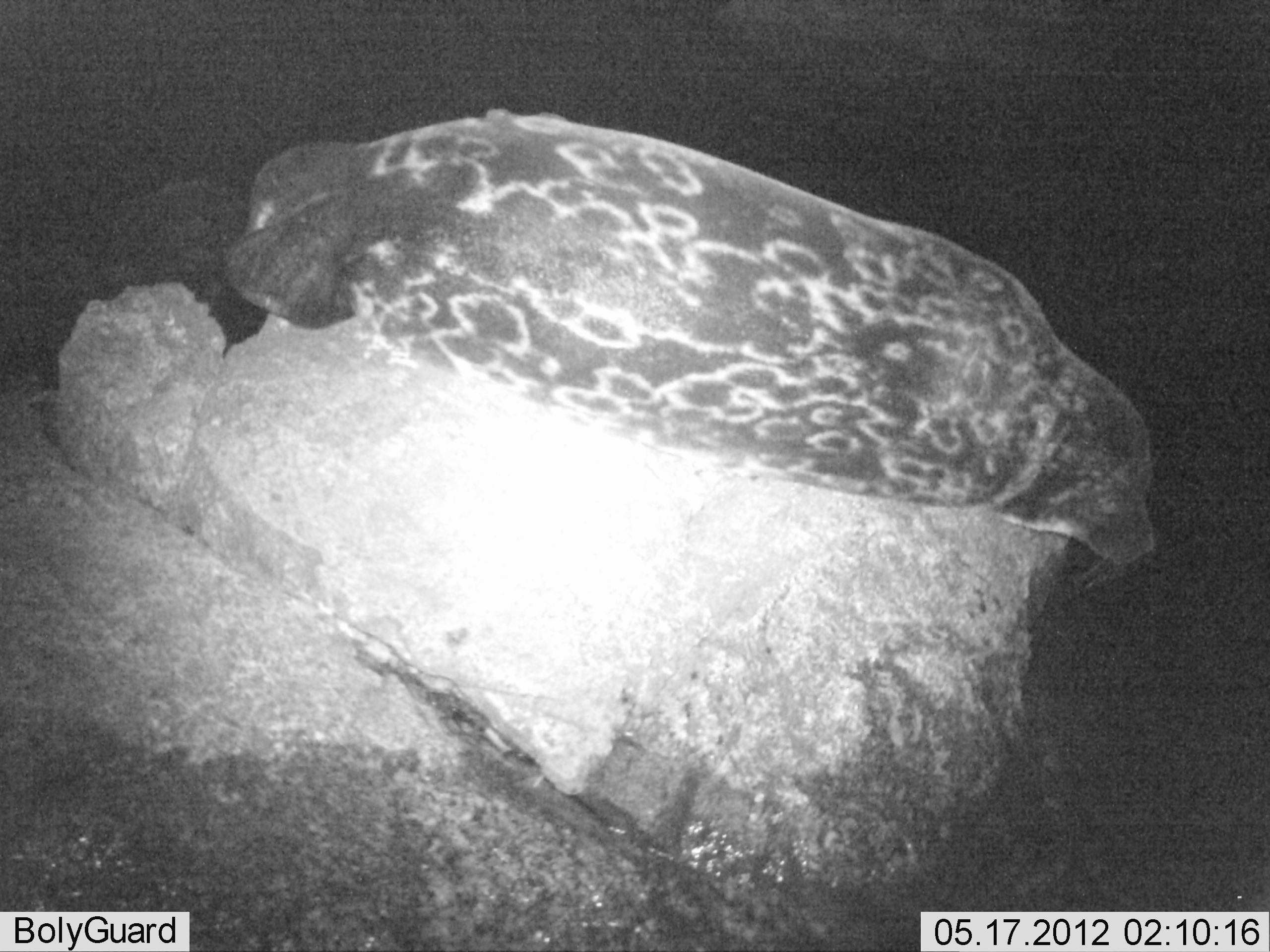}
		\endminipage\hfill
		\minipage{0.30\linewidth}
		\includegraphics[width=\linewidth]{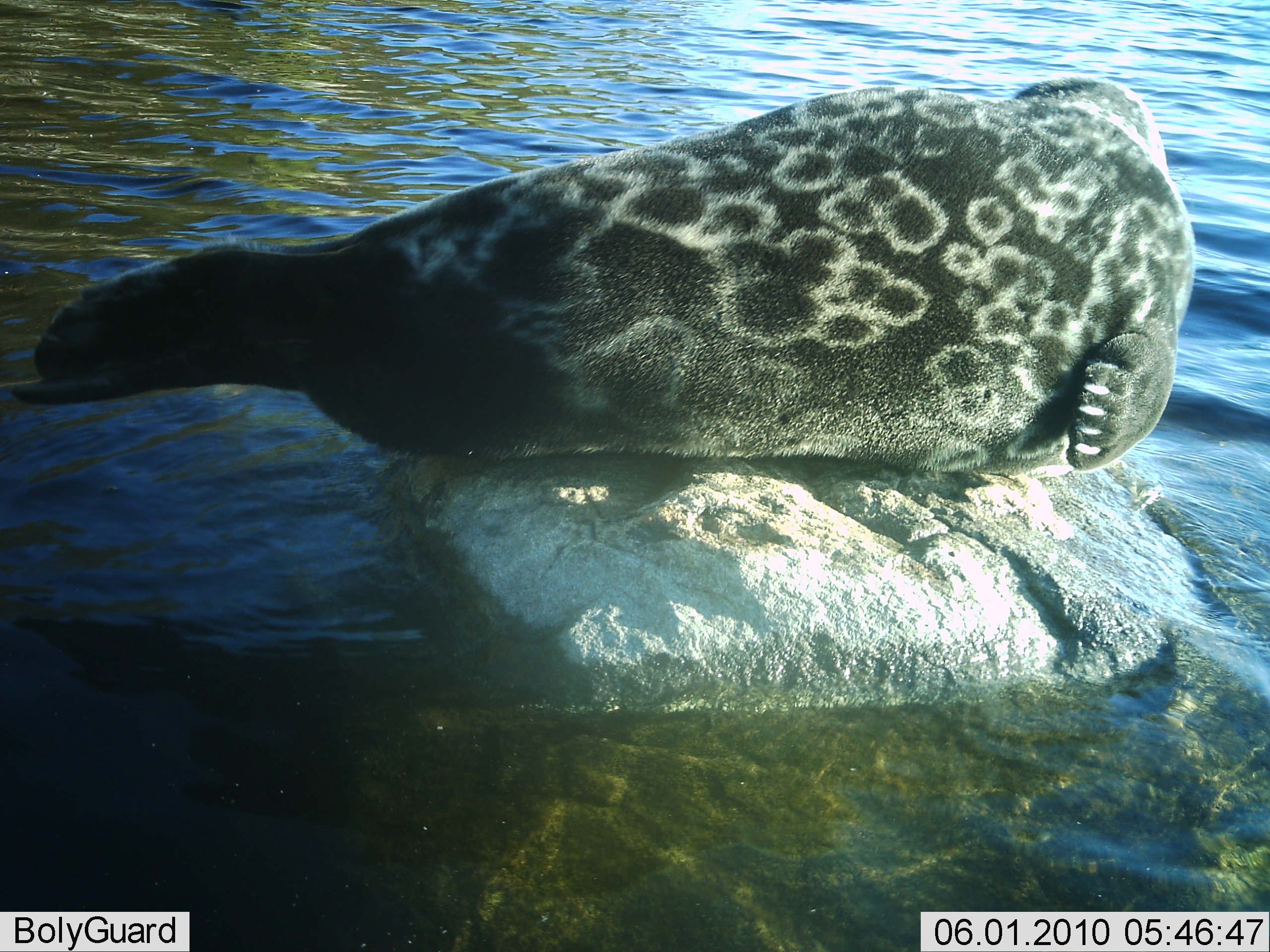}
		\endminipage\hfill
		\minipage{0.30\linewidth}
		\endminipage
		\caption{Examples of the Saimaa ringed seal images.}
		\label{fig:seal_img}
	\end{figure}

The Saimaa ringed seal (\textit{Pusa hispida saimensis}) is an endangered subspecies of the ringed seals only found in Lake Saimaa in Finland. Due to the small population size and restricted habitat, conservation efforts including monitoring are essential. Recently, efforts towards establishing photo-identification for the Saimaa ringed seals have been carried out. Preliminary studies utilizing manual identification~\cite{koivuniemi2016photo,kunnasranta2021sealed} have already provided novel information about the spatial movement and social behaviour of the animal. However, the full potential of the photo-identification is still to be realised due to the lack of automatic methods. The Saimaa ringed seals have pelage patterns (see Fig.~\ref{fig:seal_img}) that are unique to each individual and remain constant during the whole lifespan. This makes it possible to develop methods for the identification of individuals.

Various methods for automatic re-identification of animals exist~\cite{schneider2019similarity,moskvyak2020learning,cheng2020detection}. However, the low contrast between the pattern and the rest of the pelage, large pose variations, and low image quality of game cameras make the re-identification of the Saimaa ringed seals considerably more challenging than for animals (e.g., Zebra and Amur Tiger) considered in the earlier studies. This renders the existing methods ill-suited for ringed seals. Some efforts towards automatic re-identification of the Saimaa ringed seals have been made~\cite{zhelezniakov2015segmentation,chehrsimin2017automatic,nepovinnykh2018identification,nepovinnykh2020siamese}. In~\cite{chehrsimin2017automatic} various preprocessing steps to make the pattern more visible were tested together with existing species-agnostic re-identification methods. However, the accuracy remained at the unsatisfactory level. In~\cite{nepovinnykh2018identification}, the re-identification was formulated as a classification problem and was solved using transfer learning. While the accuracy was high on the used test set, the method is only able to reliably perform the re-identification if there is a large set of examples for each individual which heavily limits its usability. In~\cite{nepovinnykh2020siamese}, a 
metric learning based method for the re-identification was proposed. The method consists of the Siamese triplet network based pattern patch matching and the topology-preserving projection based comparison of query patches with the database of known individuals. The method allows one-shot identification, i.e., it is technically possible to re-identify the individual in a query image based on just one example image in the database of known individuals. However, the challenging nature of the ringed seal pattern matching limits the accuracy of the method for a wider use in practice.

In this paper, the Saimaa ringed seal pattern matching is considered. The method from~\cite{nepovinnykh2020siamese} is further developed by replacing the triplet loss with the ArcFace loss~\cite{deng2019arcface} to avoid the challenging triplet mining step and to obtain better individual separability. Moreover, the topology-preserving projection based comparison is replaced with global pooling, allowing the aggregation of the local pattern features resulting in embedding vectors that contain global features. The main contribution of the work is a novel global pooling technique that incorporates spatial distribution of features. This is obtained by perceiving feature maps as 2D probability mass functions. We then make use of eigen decomposition of covariances to represent them in the compact manner. 
We call this method as EDEN - Eigen DEcompositioN of covariance matrices of location depicting random variables. In the experimental part of the work, we show that the proposed pooling technique outperforms the baseline in pattern matching accuracy.



%
%
\section{Related Work}
%
%
\subsection{Metric Learning}
%

Training a machine learning method to re-identify animal individuals can be seen as a metric learning problem where similarity metric that quantifies how likely it is that animals in two images are the same individual is learned based on data. When combined with Convolutional Neural Networks (CNNs) based feature extraction it should be possible to learn image features that produce such a similarity metric in an end-to-end manner.

One of the most popular approaches for metric learning is to utilize the triplet loss~\cite{hoffer2015deep}. The idea is straightforward. Given three images: the anchor $x_a$, the positive $x_p$ (e.g. the same individual as anchor) and the negative $x_n$ (different individual), and some predefined margin $m$, the loss punishes the network if the distance, usually Euclidean, between the negative and the anchor examples is less than the distance between the positive and the anchor examples by $m$. Such an approach, even though providing a flexible training process, contains its own drawback, namely, triplet mining. In each iteration of the training the image triplet should be selected which is not straightforward, since to select the hardest triplet we have to calculate distances for all the possible triplets and then choose the suitable one.

To overcome the triplet mining problem, ArcFace~\cite{deng2019arcface} was proposed. The major benefits of the approach include the lack of need for triplet mining, as well as, better class separability due to the use of the angle distance instead of the Euclidean one.  ArcFace is a softmax-based loss function originally proposed for face recognition. The main idea is to approach the weights of the last fully-connected layer, located before softmax, as class centers. Then, normalizing and multiplying weights and feature vectors by some predefined value $s$ makes it possible to distribute embeddings on the hypersphere of radius $s$ and utilize geodesic distance. The loss can be formulated as:
\begin{equation}
    L=-\frac{1}{N}\sum_{i=1}^N\log\frac{e^{s(\cos(\theta_{y_i}+m))}}
    {e^{s(\cos(\theta_{y_i}+m))} + \sum_{j=1, j \neq y_i}^n e^{s\cos\theta_j}},
\end{equation}
where $s$ is a radius of the hypersphere, $\theta_{y_i}$ is an angle between predicted embedding and the vector of weights of the right class $y_i$ from the final layer, $\theta_j$ is an angle between $j$ class weights vector and the predicted by the network embedding vector, $N$ and $n$ are the batch size and number of classes (e.g., individuals) respectively, and $m$ stands for predefined additive margin, used to increase interclass separability, while decreasing intraclass variation. This results in a loss function which train the network in a simple manner while providing state-of-the-art results.

\subsection{Global Pooling}
%
CNN based metric learning provides a tool to compute similarities between images or parts of images. In order to build a full re-identification framework the correct (the most similar) individual needs to be found from the database of known individuals. This can be seen as an image retrieval problem~\cite{chen2021deep}. 
A common approach to utilize CNN features in image retrieval is to use global pooling. The idea is to spatially aggregate local features to get the embedding vector of the fixed size that incorporates global features. As all the global pooling techniques are differentiable, this makes it possible to train the model in an end-to-end manner. The major benefits also consist of rotation and translation invariant properties of the model as opposed to the use of traditional fully-connected layers as a final step~\cite{tolias2015particular}. 

Let us consider a 3D tensor with no batch dimension for brevity, computed by the last convolutional layer of the network. Its size is $C\times H\times W$ where $C$ is a number of channels, or feature maps, and $H$ with $W$ standing for the height and the width of each feature map, respectively. Let $\mathcal{X}_k$ be a set of elements of the feature map $k$ where $k\in \{1\dots C\}$. Then, the embedding vector, aggregated through global pooling, can be written as:
\begin{equation}
    \textbf{v}=\left [ v_1\dots v_k\dots v_C \right ]^\top,
    \quad v_k=P(\mathcal{X}_k),\quad k\in\{1\dots C\},
\end{equation}
where the pooling function $P$ operates on all of the elements of the 2D matrix of the feature map. This $P$ could be any aggregating function. For example, in MAC~\cite{tolias2015particular} maximum pooling was applied in the following manner:
\begin{equation}
    \textbf{v}=\left [ v_1\dots v_k\dots v_C \right ]^\top,
    \quad v_k=\max_{x\in\mathcal{X}_k}x,\quad k\in\{1\dots C\}
\end{equation}
Radenovi\'c $\textit{et al.}$~\cite{radenovic2018fine} proposed to use the generalized mean (GeM) defined as:
\begin{equation}
    \textbf{v}=\left [ v_1\dots v_k\dots v_C \right ]^\top,
    \quad v_k=\left ( \frac{1}{\left | \mathcal{X}_k \right |} \sum_{x\in \mathcal{X}_k} x^{p_k}\right ) ^ {\frac{1}{p_k}},\quad k\in\{1\dots C\}
\end{equation}
The greater the power parameter $p_k$, the more the network values strong features. With $p_k \to \infty$ it becomes a maximum function. One of the major benefits is that $p_k$ is also learnable so it can be optimized during the learning process.

%
%
\section{Proposed Method}

\subsection{Saimaa Ringed Seal Re-Identification Pipeline}

The main underlying idea of the Saimaa ringed seal re-identification is the fact that each individual seal has the unique pelage pattern. Therefore, the algorithm is based on the analysis of the pattern. 
The whole pipeline consists of the following five steps:
\begin{enumerate}
    \item Seal segmentation.
    \item Pelage pattern extraction.
    \item Patch extraction.
    \item Generation of patch descriptors.
    \item Generation of full image descriptors.
\end{enumerate}

\begin{figure}[ht]
	\begin{center}
		\includegraphics[width=1\linewidth]{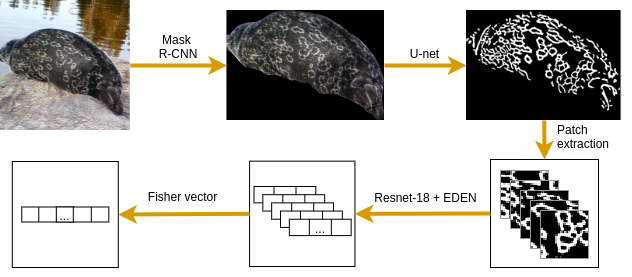}
		\caption{The whole re-identification framework.}
		\label{fig:framework}
	\end{center}
\end{figure}

The schematic of the framework is also presented in Fig.~\ref{fig:framework}.

The first step is to detect the seal and remove the background by segmenting the seal. This is done by using the Mask R-CNN~\cite{he2017mask}. The segmented and cropped seal image is then fed into the pattern extraction algorithm~\cite{zavialkin2020cnn} utilizing the U-net architecture~\cite{ronneberger2015u}. The output is a black and white image of the pattern itself. This is since it is essential to remove unnecessary information that could have negative effect on the training or the re-identification such as illumination and body parts that do not contain the pattern. Next, the pattern is cut into overlapping square patches. The idea is that different patches contain different parts of the pattern due to the differing viewpoints and seal poses, and therefore, by processing patches we are able to match specific subregions of the pattern. Each pattern patch is fed into a network to produce the feature descriptors. The descriptor network is trained by means of the ArcFace loss in order to learn the metric to distinguish different image patches. After the last convolutional layer, the global pooling layer is utilized to ensure that the final embedding is rotation and translation invariant. The development of the novel global pooling strategy is the focus of this study and is described in more detail in Section~\ref{subsec:eden}. Then, to calculate accuracy of the network, the class for the query image patch is set the same as of the closest patch from the database. 
Finally, the patch descriptors are aggregated using the Fisher vector algorithm to generate the descriptor for the whole image. The re-identification is then done by matching it to the database of known individuals by finding the image with the minimum distance to that descriptor.

\subsection{EDEN Pooling}
\label{subsec:eden}
%
The results on GeM have showed that pooling methods could be further improved by means of generalization~\cite{radenovic2018fine}. However, most of the existing global pooling techniques treat elements of feature maps as activations of neurons. For example, the idea behind MAC is that each $\mathcal{X}_k$ represents a spatially distributed unique feature. Therefore, the stronger the activation of such map, e.g. the greater the value of the element $x\in\mathcal{X}_k$, the more likely that this feature is present on the image. This view on the problem was also depicted in various other works which propose global pooling techniques, such as sum pooling (SPoC)~\cite{babenko2015aggregating}, weighted sum pooling~\cite{kalantidis2016cross}, and mixed approaches~\cite{jun2019combination,mousavian2015deep}. However, the main problem with such methods is that information about spatial feature distribution is lost. As it was mentioned before, each feature map represents some unique feature so each element of it shows how strong the feature is on the image in the given location. However, we may have several instances of the same feature on the image. Therefore, by pooling with methods like MAC or GeM we may lose information about feature distribution, e.g., the amount of strong features presented on the image.

We propose a global pooling technique that incorporates spatial distribution of features. Let us denote an activation of the feature map $k$ in the location $\left( y, z \right)$ as $\mathcal{X}_k\left( y, z \right)$ where $y \in Y_k=\{1\dots W\}$, $z \in Z_k=\{1\dots H\}$. We consider each element of the feature map as the probability of the feature appearing in that location. This results in a 2D discrete probability mass function of two random variables $Y_k$ and $Z_k$. It can be obtained from $\mathcal{X}_k$ by applying softmax to each element as 
\begin{equation}
    S(\mathcal{X}_k(y, z))=\frac{e^{\mathcal{X}_k(y, z)}}{\sum_{x\in\mathcal{X}_k}e^{x}}, 
    \quad y\in Y_k, z \in Z_k, k \in \{1 \dots C\}.
\end{equation}

\begin{figure}[ht]
	\begin{center}
		\includegraphics[width=0.8\linewidth]{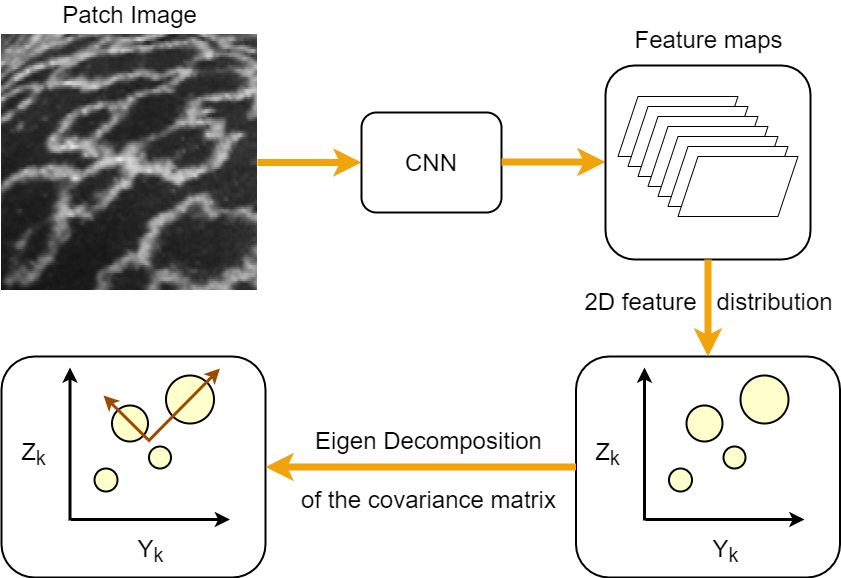}
		\caption{A pipeline of the EDEN.}
		\label{fig:eden_pipe}
	\end{center}
\end{figure}

The resulting matrix can then be compactly characterized by the mean $\mu$ and the covariance matrix $\Sigma$ of $Y_k$ and $Z_k$. However, $\mu$ and $\Sigma$ are not rotation invariant features. To overcome this, we compute eigen decomposition of the covariance matrix. The main reasoning behind this, is that eigenvalues depict normalized with respect to rotation and translation variances of the distribution. As the final step, the mean $\mu$ and eigenvectors are omitted which leaves us with two values per feature map. We call this method as EDEN where the abbreviation comes from Eigen DEcompositioN of covariance matrices of location depicting random variables. The full pipeline is shown in Fig.~\ref{fig:eden_pipe}.

%
%
\section{Experiments}
\begin{figure}[t!]
	\begin{center}
		\includegraphics[width=1\linewidth]{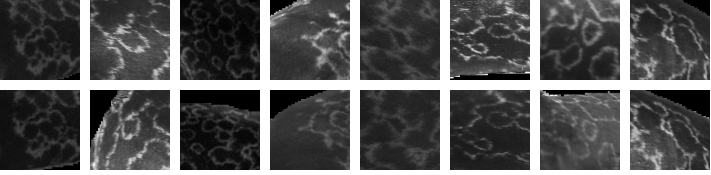}
		\caption{Examples of pattern images (patches). The patches in the second row match the patches in the first row.}
		\label{fig:original_patches}
	\end{center}
\end{figure}

All the experiments for the patch matching step were carried out on the Saimaa ringed seals dataset of image patches (see Fig. \ref{fig:original_patches}). The dataset contains, in total, 4599 images (patches of the size $256 \times 256$ pixels). The data were divided into the training set, containing 3016 images and 16 classes and the testing set, containing 1583 images and 26 classes that are different from the training classes in the training set. Each class corresponds to one manually selected location in the pelage pattern of one seal, and each sample from one class was extracted from different images of the same seal. To compute the accuracy of the method, the testing set was further divided into the database and query parts with the ratio of 1 to 2.

The dataset for the full image matching contains 2080 images and 57 individuals. The images are divided into the query and database subsets, containing 1650 and 430 images respectively. It should be noted that the network training was performed on the patches dataset and the full images were used only for testing.

ResNet-18~\cite{he2016deep} was used as a backbone network. However, all the pooling layers were replaced with SoftPool~\cite{stergiou2021refining} as it retains more information compared to originally used average pooling and is shown to improve performance. Moreover, a second order attention module~\cite{ng2020solar} was added after the 3rd and 4th blocks of the ResNet. The final global average pooling of the ResNet was then replaced by one of the considered global pooling layers and followed by the fully-connected layer with the output size of 512 and the $L^2$ normalization layer. In case of the concatenated version of the pooling layer, feature vectors were concatenated into one and then passed to the fully-connected layer as usual. For the training process, the AMSGrad~\cite{reddi2019convergence} version of the AdamW~\cite{loshchilov2017decoupled} optimizer was used. The weight decay was set to $10^{-3}$, the initial learning rate was $3 \times 10^{-5}$, and the batch size was $16$. Each network was trained for 50 epochs with the learning rate decreasing to $4.5 \times 10^{-6}$ after 30 epochs. 

\begin{table}[b!]
	\caption{Accuracy of each method used for pattern matching. FC stands for fully connected layer. EDEN + MAC and EDEN + GeM stands for concatenated features from EDEN and MAC, GeM. The best model is highlighted.}
	\begin{center}
		\begin{tabular}{|c|c|c|}
			\hline
			Method & \quad Patch matching \quad & \quad Full image matching \quad \\
			\hline
			FC & 76.11\% & 31.82\% \\
			\hline
			MAC & 84.36\% & 35.03\% \\
			\hline
			GeM & 85.97\%  & 35.64\% \\
			\hline
			EDEN & 86.26\% & 36.73\%  \\
			\hline
			\quad EDEN + MAC \quad & \textbf{86.54\%} & \textbf{37.09\%} \\
			\hline
			\quad EDEN + GeM \quad & 86.07\% & 35.64\% \\
			\hline
		\end{tabular}
		\label{tab:patch_results}
	\end{center}
\end{table}

The results of the experiments are shown in Table~\ref{tab:patch_results}.  
The accuracy is computed by comparing the predicted class with the ground truth. For the full image matching, the descriptors for patches are aggregated into a Fisher vector  which is then used to compute the accuracy in a similar manner. The results show that EDEN outperforms conventional methods such as GeM and MAC for the patch matching. Moreover, it boosts the performance of the Fisher vector significantly. However, the best results were obtained with a concatenated version of EDEN and MAC.  This is most likely due to the nature of EDEN as it only incorporates the distribution of features and not their strength, while MAC can be seen as a mode of the discrete probability distribution. Therefore, it is worth combining the both approaches.

%

\section{Conclusion}

In this paper, a challenging problem of the Saimaa ringed seal re-identification was considered. We focused mainly on the pelage pattern matching step, forming the basis for the re-identification. The method was based on the metric learning framework. Instead of the commonly used Triplet loss, the ArcFace loss was used in the training to overcome the triplet mining problem. In addition, global pooling layers were studied to address their shortcomings, especially the lack of information about feature distribution. Based on this, a novel global pooling method called EDEN was proposed. The method considers feature maps as 2D probability mass functions and performs an eigen decomposition of covariances to obtain a compact feature representation. The proposed method was shown to outperform earlier methods including GeM and MAC. However, the best performance was obtained by pairing EDEN with MAC. 
The future research include incorporating further analysis of distribution-based pooling techniques such as consideration of cross-channel feature distributions.

\section*{Acknowledgements}
The authors would like to thank the CoExist project (Project ID: KS1549) funded by the European Union, the Russian Federation and the Republic of Finland via The South-East Finland--Russia CBC 2014-2020 programme, especially Vincent Biard, Marja Niemi, and Mervi Kunnasranta from Department of Environmental and Biological Sciences at University of Eastern Finland (UEF) for providing the data and their expert knowledge for identifying the individuals.

%
%
\bibliographystyle{spmpsci}
\bibliography{reference}

\begin{thebibliography}{10}
\providecommand{\url}[1]{{#1}}
\providecommand{\urlprefix}{URL }
\expandafter\ifx\csname urlstyle\endcsname\relax
  \providecommand{\doi}[1]{DOI~\discretionary{}{}{}#1}\else
  \providecommand{\doi}{DOI~\discretionary{}{}{}\begingroup
  \urlstyle{rm}\Url}\fi

\bibitem{babenko2015aggregating}
Babenko, A., Lempitsky, V.: Aggregating local deep features for image
  retrieval.
\newblock In: ICCV, pp. 1269--1277 (2015)

\bibitem{chehrsimin2017automatic}
Chehrsimin, T., Eerola, T., Koivuniemi, M., Auttila, M., Lev{\"a}nen, R.,
  Niemi, M., Kunnasranta, M., K{\"a}lvi{\"a}inen, H.: Automatic individual
  identification of {S}aimaa ringed seals.
\newblock IET Computer Vision \textbf{12}(2), 146--152 (2018)

\bibitem{chen2021deep}
Chen, W., Liu, Y., Wang, W., Bakker, E., Georgiou, T., Fieguth, P., Liu, L.,
  Lew, M.S.: Deep image retrieval: A survey.
\newblock arXiv preprint arXiv:2101.11282  (2021)

\bibitem{cheng2020detection}
Cheng, X., Zhu, J., Zhang, N., Wang, Q., Zhao, Q.: Detection features as
  attention (defat): A keypoint-free approach to amur tiger re-identification.
\newblock In: ICIP, pp. 2231--2235 (2020)

\bibitem{deng2019arcface}
Deng, J., Guo, J., Xue, N., Zafeiriou, S.: Arcface: Additive angular margin
  loss for deep face recognition.
\newblock In: CVPR, pp. 4690--4699 (2019)

\bibitem{he2017mask}
He, K., Gkioxari, G., Doll{\'{a}}r, P., Girshick, R.B.: Mask {R-CNN}.
\newblock In: ICCV, pp. 2980--2988 (2017)

\bibitem{he2016deep}
He, K., Zhang, X., Ren, S., Sun, J.: Deep residual learning for image
  recognition.
\newblock In: CVPR, pp. 770--778 (2016)

\bibitem{hoffer2015deep}
Hoffer, E., Ailon, N.: Deep metric learning using triplet network.
\newblock In: International Workshop on Similarity-based Pattern Recognition,
  pp. 84--92. Springer (2015)

\bibitem{jun2019combination}
Jun, H., Ko, B., Kim, Y., Kim, I., Kim, J.: Combination of multiple global
  descriptors for image retrieval.
\newblock arXiv preprint arXiv:1903.10663  (2019)

\bibitem{kalantidis2016cross}
Kalantidis, Y., Mellina, C., Osindero, S.: Cross-dimensional weighting for
  aggregated deep convolutional features.
\newblock In: ECCV, pp. 685--701. Springer (2016)

\bibitem{koivuniemi2016photo}
Koivuniemi, M., Auttila, M., Niemi, M., Lev{\"a}nen, R., Kunnasranta, M.:
  Photo-id as a tool for studying and monitoring the endangered {S}aimaa ringed
  seal.
\newblock Endangered Species Research \textbf{30}, 29--36 (2016)

\bibitem{kunnasranta2021sealed}
Kunnasranta, M., Niemi, M., Auttila, M., Valtonen, M., Kammonen, J., Nyman, T.:
  Sealed in a lake—biology and conservation of the endangered {S}aimaa ringed
  seal: A review.
\newblock Biological Conservation \textbf{253}, 108,908 (2021)

\bibitem{loshchilov2017decoupled}
Loshchilov, I., Hutter, F.: Decoupled weight decay regularization.
\newblock arXiv preprint arXiv:1711.05101  (2017)

\bibitem{moskvyak2020learning}
Moskvyak, O., Maire, F., Dayoub, F., Baktashmotlagh, M.: Learning landmark
  guided embeddings for animal re-identification.
\newblock In: WACVW, pp. 12--19 (2020)

\bibitem{mousavian2015deep}
Mousavian, A., Kosecka, J.: Deep convolutional features for image based
  retrieval and scene categorization.
\newblock arXiv preprint arXiv:1509.06033  (2015)

\bibitem{nepovinnykh2020siamese}
Nepovinnykh, E., Eerola, T., K\"alvi\"ainen, H.: Siamese network based pelage
  pattern matching for ringed seal re-identification.
\newblock In: WACVW (2020)

\bibitem{nepovinnykh2018identification}
Nepovinnykh, E., Eerola, T., K{\"a}lvi{\"a}inen, H., Radchenko, G.:
  Identification of {S}aimaa ringed seal individuals using transfer learning.
\newblock In: ACIVS, pp. 211--222. Springer (2018)

\bibitem{ng2020solar}
Ng, T., Balntas, V., Tian, Y., Mikolajczyk, K.: Solar: second-order loss and
  attention for image retrieval.
\newblock In: ECCV, pp. 253--270. Springer (2020)

\bibitem{radenovic2018fine}
Radenovi{\'c}, F., Tolias, G., Chum, O.: Fine-tuning cnn image retrieval with
  no human annotation.
\newblock TPAMI \textbf{41}(7), 1655--1668 (2018)

\bibitem{reddi2019convergence}
Reddi, S.J., Kale, S., Kumar, S.: On the convergence of adam and beyond.
\newblock arXiv preprint arXiv:1904.09237  (2019)

\bibitem{ronneberger2015u}
Ronneberger, O., Fischer, P., Brox, T.: U-net: Convolutional networks for
  biomedical image segmentation.
\newblock In: MICCAI, pp. 234--241 (2015)

\bibitem{schneider2019similarity}
Schneider, S., Taylor, G.W., Linquist, S., Kremer, S.C.: Similarity learning
  networks for animal individual re-identification-beyond the capabilities of a
  human observer.
\newblock WACVW  (2019)

\bibitem{stergiou2021refining}
Stergiou, A., Poppe, R., Kalliatakis, G.: Refining activation downsampling with
  softpool.
\newblock arXiv preprint arXiv:2101.00440  (2021)

\bibitem{tolias2015particular}
Tolias, G., Sicre, R., J{\'e}gou, H.: Particular object retrieval with integral
  max-pooling of cnn activations.
\newblock arXiv preprint arXiv:1511.05879  (2015)

\bibitem{zavialkin2020cnn}
Zavialkin, D.: {CNN}-based ringed seal pelage pattern extraction.
\newblock Master's thesis, Lappeenranta-Lahti University of Technology LUT,
  Finland (2020)

\bibitem{zhelezniakov2015segmentation}
Zhelezniakov, A., Eerola, T., Koivuniemi, M., Auttila, M., Lev{\"a}nen, R.,
  Niemi, M., Kunnasranta, M., K{\"a}lvi{\"a}inen, H.: Segmentation of {S}aimaa
  ringed seals for identification purposes.
\newblock In: ISVC, pp. 227--236. Las Vegas, USA (2015)

\end{thebibliography}
\addcontentsline{toc}{section}{REFERENCES}

\end{document}